\documentclass[seceq]{iis}

\articleno{20YY.Z.NN}
\papertype{SHORT COMMUNICATION}
\subjclass{2010 Mathematics Subject Classification: 
Primary XXXXX, Secondary XXXXX, etc.}
\title{Mobile Augmented Reality Framework with Fusional Localization and Pose Estimation}
\runtitle{The short title in the header}

\author{
  \name{Songlin \surname{HOU}}$^{1}$\footnotemark[2]\thanks{Corresponding author. E-mail: shou@wpi.edu},
  \name{Fangzhou \surname{LIN}}$^{2,3}$\thanks{Equal Contribution as Co-First Authors},
  \name{Yunmei \surname{HUANG}}$^{4}$\footnotemark[2],
  \name{Zhe \surname{PENG}}$^{5}$,
  \name{Bin \surname{XIAO}}$^{5}$
}

\inst{$^{1}$Department of Computer Science, Worcester Polytechnic Institute, \address{Worcester 01609, USA}\\
$^{2}$Department of Robotics Engineering, Worcester Polytechnic Institute, \address{Worcester 01609, USA}\\
$^{3}$Unprecedented-scale Data Analytics Center, Tohoku University, \address{Sendai 980-8577, Japan}\\
$^{4}$Department of Forestry and Natural Resources, Purdue University, \address{West Lafayette 47907, USA}\\
$^{5}$Department of Computing, The Hong Kong Polytechnic University, \address{Hong Kong 999-077, China}}

\runauthor{The author(s)' name(s) in the header}

\support{This work is supported by Grant XYZ.}

\recdate{20YY}{1}{DD}%% set by the editors
\accdate{20YY}{2}{DD}%% set by the editors
\pubdate{20YY}{3}{DD}

\abst{As a novel way of presenting information, augmented reality (AR) enables people to interact with the physical world in a direct and intuitive way. 
While there are some mobile AR products implemented with specific hardware at a high cost, the software approaches of AR implementation on mobile platforms(such as smartphones, tablet PC, etc.) are still far from practical use. GPS-based mobile AR systems usually perform poorly due to the inaccurate positioning in the indoor environment. Previous vision-based pose estimation methods need to continuously track predefined markers within a short distance, which greatly degrade user experience. This paper first conducts a comprehensive study of the state-of-the-art AR and localization systems on mobile platforms. Then, we propose an effective indoor mobile AR framework. In the framework, a fusional localization method and a new pose estimation implementation are developed to increase the overall matching rate and thus improving AR display accuracy. Experiments show that our framework has higher performance than approaches purely based on images or Wi-Fi signals. We achieve low average error distances (0.61-0.81m) and accurate matching rates (77\%-82\%) when the average sampling grid length is set to 0.5m.}

\kword{\kw{augmented reality}, \kw{indoor localization}, \kw{fusional localization}, \kw{location based service}}

\begin{document}
\maketitle

\section{Introduction}
% no \IEEEPARstart

Wireless techniques are extensively utilized in various mobile applications, significantly driving the rapid expansion of personal use services such as location-based services (LBS). As a subset of promising LBS applications, indoor mobile augmented reality (AR) has been attracting considerable attention in personal use, facilitated by advancements in the computational capabilities of mobile devices. While most existing AR applications on the market rely on vision tracking, inertial tracking, or GPS tracking\cite{billinghurst2015survey}, implementing indoor mobile AR is often considered challenging due to 1) the difficulty of achieving accurate user localization in indoor environments with minimal or no GPS signals\cite{angelin_gladston__2018} and 2) the limitations of sensor types available in standard mobile devices or specialized settings which are not common in most mobile phones, such as point cloud technologies\cite{vishva_nitin_patel__2021}. Furthermore, while coarse-grained localization may suffice for outdoor environments, it is often inadequate for indoor scenarios, where the intricate structures and confined spaces of indoor environments pose additional challenges.

In indoor environments, a mobile device needs to offer the most probable position before providing augmented reality (AR) to users by predicting or matching based on sensor inputs. Thus, for an indoor mobile AR application, indoor localization and pose estimation are two key components in implementation. 

\subsection{Indoor Localization Techniques}
% Background of indoor localization
Many indoor localization approaches\cite{s30,s10,s5,s7,niu2015wicloc,s24} based on different mobile phone sensors have been proposed to facilitate indoor AR, such as wireless local area network (WLAN), Bluetooth, inertial sensors, geomagnetism, vision analysis or audible sound based systems. With most methods using vision analysis adopting different fundamental techniques from the others, we categorize indoor localization methods as non-vision-based methods and vision-based methods. As for the vision-based methods, we use image matching and image retrieval interchangeably with no discrimination. Several sensors used in indoor localization such as radio-frequency identification (RFID), infrared are not discussed in this article since the extra customized hardware is often needed other than mobile phones. 

% Intro. of related work in indoor localization
For indoor localization, non-vision-based approaches with inertia sensors and WLAN are most widely studied for its low cost both on users and infrastructure deployment. For inertia sensors\cite{s24,jianguo_sun__2023} like gyroscope and accelerator, relative movements are measured and accumulated in order to calculate user's moving trajectory and estimate the position. Localization methods\cite{niu2015wicloc,s23,s10} using WLAN are usually based on trilateration or fingerprinting. Trilateration-based methods\cite{cheng2018fast} largely depend on the signal-strength-to-distance relationship with generally lower precision than methods using fingerprinting, while a large amount of sampling work is usually required by fingerprinting. The combination of WLAN and inertia sensors to improve precision can also be found in many studies\cite{s24}. With the booming of machine learning and deep learning, many learning-based methods are also proposed for this task. Sebetic et al.\cite{omer_gokalp_serbetci__2024} provided a wireless channel-aware data augmentation method that can significantly improve localization accuracy with reduced measurement data. Some researchers\cite{yong_deng__2024, rafael_alves_de_aguiar__2024} also applied LSTM on wireless signals to enhance generalization and reliability.

On the other hand, vision-based indoor localization uses image pattern recognition and matching (or retrieval) to deal with AR-based indoors localization problem. The methods can be categorized into marker-based localization, semantics-based localization, and photo matching. Marker-based methods use pattern recognition to detect predefined markers for tracking and are now mostly used on robots\cite{lopez2016accuracy}. Semantics-based localization \cite{s29} uses semantic descriptors as features to locate users. Applications in \cite{sadeghi2017ocrapose} demonstrate how to use OCR (Optical Character Recognition) as semantic features for localization and AR. A similar pedestrian localization system called OCRAPOSE II\cite{sadeghi2017ocrapose} also locate users using OCR in detecting numbers. Feature-based localization uses image features taken from the camera to retrieve possible images of indoor environments on servers. Works from \cite{xin_lin_2023} use features extracted from images to identify locations. Apart from that, learning-based methods become popular due to their high performance exhibited in image recognition. works such as \cite{gabriel_toshio_hirokawa_higa__2024, d__chen__2024} formalized indoor localization as an image classification task. Moreover, some works use techniques usually found in vision-related tasks to process signal data. For example, Saideep et al.\cite{saideep_tiku__2023} and Ayush et al.\cite{ayush_kr__mittal__2018} use CNN to classify images generated from WLAN signals. Huiqing et al.\cite{huiqing_zhang__2023} adopt both CNN and wavelet transform to process WLAN data directly.

Apart from the two categories mainly discussed above, there are techniques that have been widely used to facilitate the task of indoor localization. Structure from Motion (SfM) reconstructs 3D environments from 2D images, which can be used as a navigable maps, setting the stage for localization. Apart from it, point cloud, which is usually generated with depth camera, is often used for map generation. Siddharth et al.\cite{siddharth_chauhan__2024} focus on the image generation with point cloud registration. Some deep learning-based methods\cite{lin2023hyperbolic, lin2024infocd} can also be used to help recover defeats commonly found in point cloud during map construction. Xinying He et al.\cite{xinying_he__2023} proposed a framework without keypoint detection and achieved enhanced performance in texture-poor scenes. Besides, Simultaneous Localization and Mapping (SLAM) enables a system to determine its position within an unknown environment while simultaneously building a map of that environment. The SLAM system is usually implemented in a way in which features collected from multiple sensors (including vision and signal data) are fused, for example, Niraj Pudasaini et al.\cite{niraj_pudasaini__2024} designed a SLAM system (SPAQ-DL-SLAM) using neural networks to enhance efficiency and accuracy in robots in resource-constrained embedded platforms. Muhammad et al.\cite{shoukat2024cognitive} utilize a graph optimization algorithm combining YOLOv5 and Wi-Fi fingerprint sequence matching, achieving satisfactory accuracy in complex environments.

% weakness of related work in indoor localization

The performance of non-vision-based indoor localization methods varies. Methods based on inertia sensors are usually hampered by the inferior hardware quality. Trilateration with WLAN\cite{cheng2018fast} can grow unsteady even in general occasions due to signal fluctuation. Fingerprinting is comparatively usable since it's less susceptible to signal changes. Vision-based indoor localization methods can provide good performance most of the time, but they share some restriction by some inevitable limitations. A proper illumination environment with minimal obstruction is needed, and the details of varieties of patterns and colors can strongly affect the precision. When these methods are used in motion, motion blur and shake of camera can degrade the precision too. With the boom of machine learning and deep learning, some models are also used for localization, like applying $K$ nearest neighbors (KNN)\cite{xie2016improved,s23}, support vector machine (SVM)\cite{chriki2017svm} on RSSI in non-vision-based methods. Or using convolutional neural network(CNN)\cite{chen2018indoor, shoukat2024cognitive} to retrieve images to help locate users. These methods turn out to be more accurate in localization; however, a drastically larger amount of data is usually required to ensure the performance, thus making it limited in some scenarios. For image matching methods using CNN, a large computational cost should be considered in real-time localization since high delay and power consumption can easily occur. Besides, we noticed while the matching methods based on CNN work generally well in outdoor environments (where landmarks are visible), they tend to generalize poorly on environment images taken indoors, especially when they are taken in motion. Moreover, pure learning-based image matching does not account for the relative distance between sampling positions and localized positions, which can introduce additional localization errors. These disadvantages suggest that learning-based image matching alone might not be the best choice for indoor localization when high precision is required.  

\subsection{Pose Estimation Techniques}
% Background of pose estimation
Pose estimation\cite{s37} is the key technique to determine the locations and orientations of virtual objects. Its implementation can also be either vision-based or non-vision-based. For most AR applications, pose estimations are vision-based. These approaches seek matches for the point correlations between the current views and the predefined patterns or 3D digital models and determine the ways of rendering in perspective modes. The non-vision-based implementations use some data sources other than visual information, such as users' localization, orientations detected using mobile phones' sensors, etc.

% Intro. of related work in pose estimation
Vision-based implementations of pose estimation can be divided into Pose from Orthography and Scaling \& Iteration (POSIT) \cite{dementhon1995model}, fiducial marker detection, and contour tracking. As a standard way to solve the PnP problem in pose estimation, POSIT  estimates the position and the rotation of the camera relative to the target through 3D coordinates of feature points from the real model, and the corresponding 2D coordinates of image points from the perspective view. Planar fiducial markers use image patterns\cite{s24,lopez2016accuracy} in the real world as a marker for detection and tracking, and usually realize pattern recognition and camera pose estimation simultaneously. Contour tracking uses the contours of a 3D digital model to match the models.

% weakness of related work in pose estimation

Vision-based pose estimation implementations share some common limitations: they can be strongly affected by the input image data. An image without proper focusing, proper illumination and occlusion can result in bad performance. Vision-based AR implementations are sensitive to distance since necessary information for recognition and tracking can only be captured when targets close to the camera. While in indoor LBS, objects to be augmented can be far from users. Traditional vision-based methods will be unsuitable for indoor uses. Non-vision-based pose estimation implementations\cite{s5,s24} have less disturbance than vision-based ones since they don't need any optical marker. However, they can also be vibration-prone and less visually appealing since the alignments and poses of visual objects are entirely dependent on the hardware qualities which are usually poor. Besides, GPS-based AR systems will fail to provide precise localization information in indoor environments.

To illustrate some classic cases, we compare the implementations of several indoor AR systems. The results are shown in Table 1.
% our work
\subsection{Our Contributions}

In order to mitigate the precision loss in current localization and pose estimation approaches and provide a more robust indoor AR implementation, we propose an indoor mobile AR framework which is deployed in android smartphones by 1)fusional localization with Wi-Fi and images as well as 2)a location-based pose estimation. In the fusional localization, distance ratio (DR) is introduced to compensate for the distance loss in image matching. We also give the implementation of location-based pose estimation based on relative Cartesian coordinate system  for AR display. An android application using the framework is developed and tested in the experiments.

We conclude our contributions as follows:

\begin{itemize}
\item We present a comprehensive review of the two key techniques in indoor mobile AR implementation, indoor localization, and AR display.
\item We explore and test the current vision-based indoor localization options and identify the common generalization issue widely present in deep learning methods (CNN) for indoor LBS applications. 
\item We propose a complete framework which fully supports indoor mobile AR with optimizing methods to improve image matching accuracy and pose estimation implementation for AR display.
\item We evaluate and compare the performance of the proposed system with and without the methods proposed in the framework. The proposed method gives lower average error distance ($0.61-0.81$ meters) than methods only based on Wi-Fi($3.81-4.18$ meters), and on SIFT\cite{lowe2004distinctive}($1.29-1.44$ meters). The reference point matching rate of the proposed method($77\%-82\%$) is much higher than methods only based on Wi-Fi($27\%-31\%$), and on SIFT($55\%-59\%$). Our method also shows a more stable performance and a higher generalization ability for CNN-based methods.
\end{itemize}

%Non-vision-based AR systems usually come with GPS. These implementations generally use users' position to determine whether virtual objects should be shown without any markers.

%However, for indoor mobile AR, these approaches tend to perform poorly since it is hard for users to observe a predefined marker (i.e. makers are invisible in most cases). Besides, it's difficult to predefine markers in every corner for continuous tracking. Finally, the near recognition distance can be a great bottleneck for indoor use for mobile devices.

% You must have at least 2 lines in the paragraph with the drop letter
% (should never be an issue)

\begin{table}
\centering
\caption{Comparasions of Several Indoor Mobile AR or Localization Systems }
\begin{tabular}{|p{2.0cm}|p{2.2cm}|p{3.0cm}|}% 通过添加 | 来表示是否需要绘制竖线
\hline  % 在表格最上方绘制横线
\textbf{Methods/Authors} & \textbf{Data Sources} & \textbf{Approaches}\\
\hline
\hline
OCRAPOSE II\cite{sadeghi2017ocrapose}&   Images  & OCR      \\
\hline
%ARQuake\cite{s12}&   Wearable PC&           Images(Marker)          \\
%\hline
%JongBae\cite{s25}&    Mobile PC&                   Images(Marker)        \\
%\hline
% CAViAR\cite{s26}&    Images&              Feature Matching\\
Yong Deng\cite{yong_deng__2024} & iBeacon(FP) & Deep Learning\\
\hline
% Remi\cite{s27} and Ahn\cite{s24}       &Image&     Feature Matching, Pose Estimation    \\
\hline
%Danqing\cite{s23}    &Smartphone&                 Wi-Fi   \\
%\hline
Taira\cite{taira2018inloc} &Images&  Dense Matching\\
\hline

Abbas\cite{abbas2019wideep} &    Wi-Fi(FP)  & Deep Learning \\
\hline

Xie\cite{xie2016improved}   & Wi-Fi(FP)  & KNN  \\
\hline

Chen\cite{chen2015fusion}  & Wi-Fi,PDR,Images & Kalman Filter \\
\hline

Chriki\cite{chriki2017svm} & Wi-Fi & SVM \\
\hline

Yang\cite{y__yang__2020} & Wi-Fi, Images & CNN, Dense Matching \\
\hline

\end{tabular}

\end{table}

\section{A New Indoor AR Framework}
\label{section3}

\begin{figure}
\centering
\includegraphics[height=3cm ,width=8.5cm]{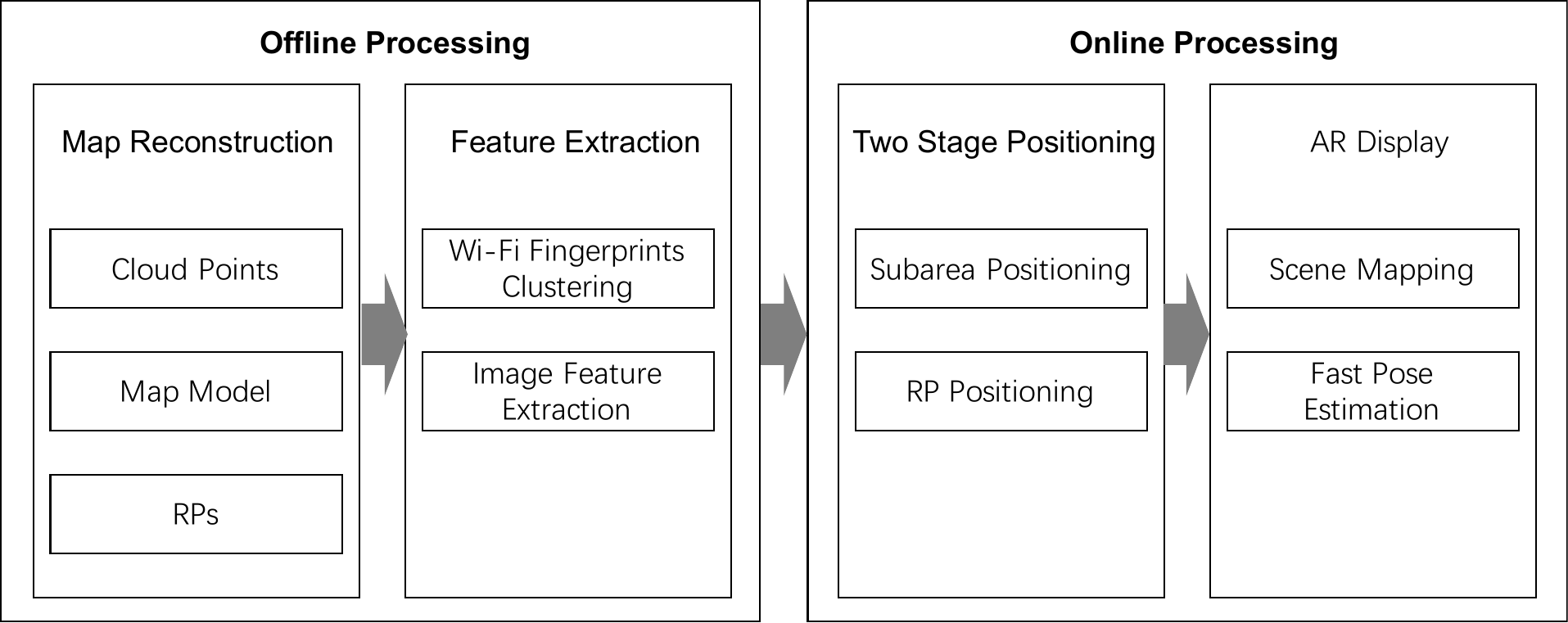}
\caption{The structure of the proposed indoor AR framework}
\label{fig_1}
\end{figure}

We propose a new indoor AR framework based on smartphone localization and pose estimation. The proposed framework provides real-time point-level AR display with accurate environment information. Users can get information about the surroundings as well as their current positions through AR information. Since the heavy computing operations (Wi-Fi and image matching) are done on the server, and smartphones simply act as the role to collect and send data, so the computation cost on the smartphones can be treated acceptable for most scenarios.

This framework consists of two stages, i.e., offline processing and online processing.
In the offline processing, we first reconstruct a 3D indoor map with various environment information, including room labels and desk numbers.
Then, in order to locate users in the indoor environment, we extract features from the data collected at each reference point in the reconstructed map.
In the online processing, we propose a new method to locate users from Wi-Fi fingerprints and images.
Moreover, a new pose estimation approach is developed to conduct the AR display of various indoor environment information.
The structure of this framework is shown in Fig. \ref{fig_1}. Next, we detail each component.

\subsection{Map Reconstruction}

An indoor map is needed before localization and AR display in our approach. Most indoor localization approaches are based on existing 2D floorplans\cite{s5} or generated floorplans\cite{peng2018indoor,peng2018crowdgis} using various sensing data. However, most of them cannot reveal structures in the vertical direction. Thus, to improve the accuracy in the AR display, we first reconstruct a 3D indoor map with various environment information in the offline processing.

\begin{figure*}[ptb]
\centering
%\includegraphics[width=13cm]{pic/figure2.eps}
%%\includegraphics[width=4cm]{pic/figure2(b).eps}
%%\includegraphics[width=4cm]{pic/figure2(c).eps}
%%(a)\hspace{1.7in}(b)
%\includegraphics[width=13cm]{pic/figure2.pdf}
\includegraphics[width=13cm]{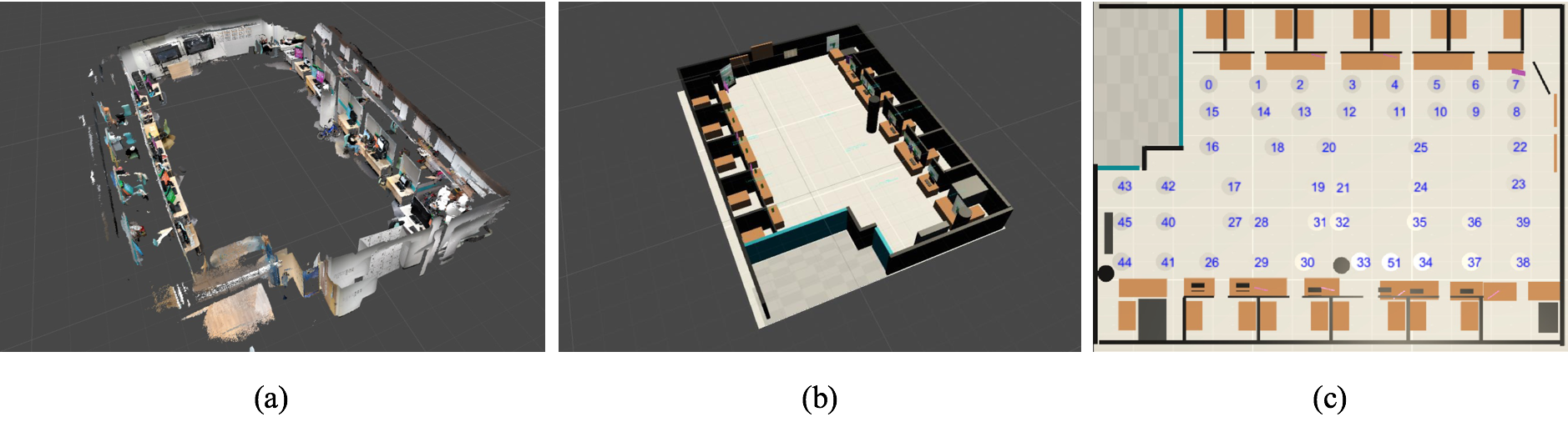}
\caption{(a) The original cloud point captured using Google's Tango device. (b) The 3D model handcrafted based on the cloud point. (c) The top-view of the 3D map with RPs annotated.}
\label{fig_2}
\end{figure*}

To reconstruct a 3D indoor map, we capture the 3D structure data using a depth camera built on Google's Tango device. Since the 3D reconstruction of a large environment can be time-consuming, we divide the whole environment into smaller areas and point clouds of these areas are collected in parallel. The point cloud of the whole environment can be assembled easily based on the overlaps of these sub-regional point clouds. The point cloud can be served as the 3D map directly, but in our case, we handcraft a 3D model using simple geometric meshes based on the point cloud to make it more user friendly. Annotated RPs (Reference Points) are marked evenly and of grid distribution with $0.5$ average meters' distance interval and confined in the walkable area in this map. The reconstruction process can be illustrated in Fig. \ref{fig_2}.

The grid distance interval is a trade-off factor in our implementation. A smaller grid length can improve localization precision in most cases, however more labor-cost is required especially when the sampling environment is large. If the AR display precision is not critical, higher sampling distances($2.0 - 10.0m$) can be used for distant AR display in spacious areas and  lower sampling distances($0.5 - 2.0m$) are only reserved for delicate near field AR display in smaller areas for labor saving.

\subsection{Feature Extraction}

Based on the reconstructed 3D indoor map, we extract features from the data collected at every reference point (RP) in two steps: Wi-Fi fingerprints clustering and image feature extraction. At every RP, we collect many Wi-Fi fingerprints and images of surrounding with fixed focal length. The fingerprints and visual features derived from images are regarded as features used in this framework.
A typical fingerprint of a RP takes the form of a set which contains signal strengths from different APs (Access Points). The fingerprint of RP $b$ is defined as equation \ref{eq:fingerprint_b} ($\bm{FP(\cdot)}$ is used to denote the fingerprint of a given RP). And $\bm{AP}^b_n$ indicates the signal strength of the $n$th AP collected at RP $b$ . 

\begin{equation}
    \bm{FP}(b) := \{\bm{AP}^b_1, \bm{AP}^b_2, \bm{AP}^b_3, ..., \bm{AP}^b_n\} 
    \label{eq:fingerprint_b}
\end{equation}

Similarly, the visual features of RP $b$ can be expressed as a set containing multiple key point collections of images taken at RP $b$. It takes the form of equation \ref{eq:vf}.

\begin{equation}
    \bm{VF}(b) := \{\bm{KP}_1, \bm{KP}_2, \bm{KP}_3, ..., \bm{KP}_n \}
    \label{eq:vf}
\end{equation}

$\bm{KP}_i$ denotes the key point collection of the $i$th image taken at RP $b$, which is defined as equation \ref{eq:kpn}. The $j$th element of the collection $\Vec{pt_j}$ is a vector contains the pixel characteristics of the $j$th key point.

\begin{equation}
    \bm{KP}_i := \{ \Vec{pt_1}, \Vec{pt_2}, \Vec{pt_3}, ..., \Vec{pt_n} \}
    \label{eq:kpn}
\end{equation}

In order to facilitate this process, we designed an Android application, which was installed on Samsung Galaxy S6, to collect fingerprints and images at the same time. At each RP, RSSs (Received Signal Strength) of different APs (access point) and photos at different viewpoints are recorded. The phone camera used for collection has 16 MP resolution (f/1.9 aperture) and 1/2.6" sensor size. The original image file collected each has 2656 x 1494 pixels (16:9 aspect ratio), and was taken without Auto HDR mode and LED flash. 
At each RP, we collected $50$ fingerprints and $50$ images; although many of the photos lack salient visual features, they are captured to truthfully and clearly depict the environment.

\textbf{Wi-Fi fingerprints clustering.}
When size of Wi-Fi fingerprints grows, matching process can soon become time-consuming. Clustering is generally used to reduce the number of fingerprint matchings and improve robustness against outliers. K-Medoids and DBSCAN are usually used to cluster fingerprints. We propose a two-step approach to cluster Wi-Fi fingerprints efficiently.

In the first step, for all the fingerprints collected at one RP, we need to find the centroid of these fingerprints. RL-clustering(combination of the authors names)\cite{s17} can be used to identify the central point. In our experiment, $45$ fingerprints are selected to represent $45$ RPs accordingly.
Then, we apply RL-clustering again on these selected central fingerprints to find $K$ subarea clusters. These $K$ subarea clusters defines $K$ subareas and we store these $K$ central points on server. We also store the list of RPs which are related to each subarea cluster. In our experiment, $K$ is set to $3$.

\textbf{Image feature extraction.}
Image features extraction is the process of getting numeric representation of prepared images. An image can contain a lot of features which can describe the content. Although human are good at differentiate various types of images, it becomes difficult for computers. Image features are commonly used for recognition. Since similar images tend to be close in features, we can retrieve similar images by comparing image features. SIFT (Scale-Invariant Feature Transform) is usually used in image retrieval for its generally higher accuracy. In SIFT, the operations on DoG (Difference of Gaussian) images can be approximated as the first order derivation in the scale space\cite{park2014robust}. To generate more key points, we can calculate the $n$th order derivations, which takes the form of equation \ref{high_order_deriv}.

\begin{equation}
    \frac{\partial^n L(x,y,\sigma)}{\partial \sigma^n} = 0
    \label{high_order_deriv}
\end{equation}

In our experiment, key points of up to $4$th order derivatives are calculated. We extract the key points from all the images from every RP and store them in the database. For the sake of brevity, we denote the SIFT key points with higher order derivatives as SIFT since they are equivalent in nature.

\textbf{Challenges with CNN-based matching} CNN is widely used for image classification. The images features generated by CNN can also be used for image matching. However, there are several challenges in terms of accurate indoor positioning.

First, it is generally impractical to train a CNN for image classification (with ID of each RP as labels) because of extensive data required for training and the self-similarity nature of indoor images. Extant vision-based approaches using CNN generally requires extensive labor for data collection and labeling, which is not applicable when images taken at different positions are too similar or when no salient features are available.

Second, extracting features with CNN for matching requires lots of additional work. Using the machine-generated features can help find the image closest to the current captured photo, but the distance difference cannot be directly measured since these features are not related to pixel locations. Taira et al.\cite{taira2018inloc} proposed a method to do pose estimation after matching images with CNN-based features, but it requires a lot more computation such as working with both low and high resolution of the query image during matching and requires multiple verification steps for pose estimation. The precision of map construction can also greatly influence the performance.

\subsection{Two Stage Positioning}
\begin{figure}[ptb]
\centering
\includegraphics[width=8.5cm]{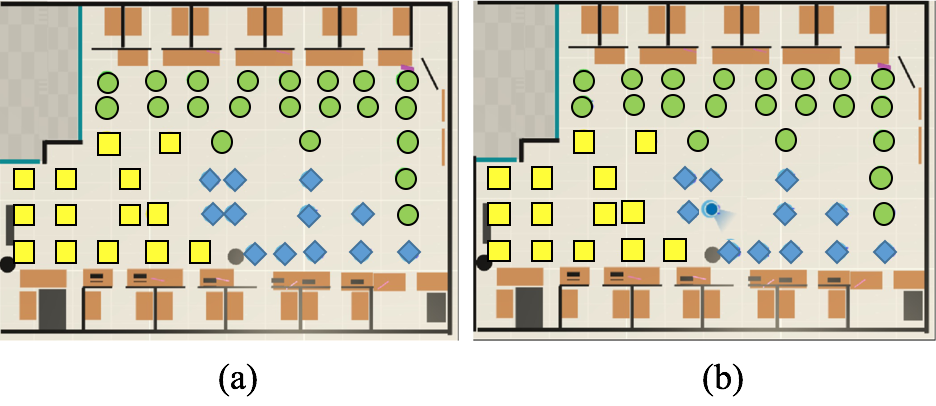}
\caption{Two steps in determining the user position. (a) subareas on maps. (b) user position.}
\label{fig_4}
\end{figure}

During the online processing, we first propose a new method to locate users from Wi-Fi fingerprints and images.
This method includes two steps: 1) subarea positioning using Wi-Fi fingerprints matching, and 2) RP positioning using image matching. The combination of Wi-Fi fingerprints and image matching is able to determine the user's current position and orientation. Compared with methods solely depending on vision, our approaches can be more time-efficient since only a small number of images are to be compared after Wi-Fi fingerprints eliminating most impossible images before matching. It is also more robust since methods depending on solely vision analysis may fail when many photos taken at different places look similar. This process is shown in Fig. \ref{fig_4}. In image (a), three subareas are created, each consisting of several RPs, which are distinguished by different colors and shapes. In image (b), the user is localized to be on one of the RP. 

\textbf{Subarea positioning.}
To determine which subarea the user is currently in, we compare the observed fingerprints captured by user's device with the premade $K$ central points of subarea clusters. Various metrics can be chosen, like Euclidean distance, cosine similarity and Manhattan distance. In our experiment, we choose cosine similarity since it gives us least error rate.

\textbf{RP positioning.}
Accurate positions can be further obtained through image matching (or retrieval) with photos taken at the RPs in this subarea when subarea positioning is done. In order to find the image most similar to the user's current scene photo, image feature extracted from the photo captured by the user's camera is compared with the image features stored in database. To calculate the similarity, the key points generated from $1-4$th order derivatives are matched.

%Key point based similarity ($KPS$) is a metric commonly used to match and find similar pairs of images. For two images with m key point matches, the feature distance between key point descriptors $P_{ni}$  and $P_{nj}$ in match $n$ is used. For two images, the $KPS$ is estimated by the following:

%\begin{equation}
%KPS=\left\{\frac{1}{m} [\sum_{n=1}^{m}dist(P_{ni},P_{nj)}]+1\right\}^{-1}.
%\label{equ_1}
%\end{equation}

There have been a lot of key point based retrieval methods \cite{park2014robust,zheng2018sift}, however, most are not directly applicable to the images with different shooting distances because they do not consider the distance diversity (which can be reflected in differences in scales and orientations, etc.) between matched images. Two images of the same environment taken at different distances can be of high similarity in these methods, but a huge loss in will occur when it is used in localization. Distance compensation is needed in this scenario as the offset for key point similarity and can leverage matching accuracy.

\textbf{Distance compensation.}
In our system, we propose a new metric, i.e., distance ratio ($\bm{DR}$), to implement distance compensation and assist matching similar images.
The $\bm{DR}$ is used to measure the proximity of a photo relative to another of the same scene. Relative to photo $p$, $\bm{DR}(q|p)  >1$ when objects in photo $q$ are closer in physical space than these objects in $p$, while $\bm{DR}(q|p)< 1$ means more distant. For a photo $A$ captured from the user's camera, the distance ratio $\bm{DR}$ of the image $B_i$ from $m$ images$(B_1,B_2,... ,B_m)$ with top keypoint similarity relative to $A$ is defined as follows:
\begin{equation}
\bm{DR}(B_{i}|A)=\frac{1}{(N-1)N}\sum_{n=1}^{N}\sum_{j=1}^{N}\frac{|B_{i}^{(n)},B_{i}^{(j)}|}{|A^{(n)},A^{(j)}|}, 1\leq i \leq m
\label{equ_2}
\end{equation}
where
\begin{equation}
|P,Q|=\sqrt{(P_{x}-Q_{x})^2+(P_{y}-Q_{y})^2}.
\label{equ_3}
\end{equation}
In (\ref{equ_2}), $B_{i}^{(n)}$ and $A^{(n)}$ are two key points in image $B_{i}$ and $A$ respectively belonging to the $n$th match between the two images, $N$ is set to 10 in our experiment($N>1$), and the 10 matches with the shortest distance between key points are chosen. In (\ref{equ_3}), $|P,Q| $function stands for the Euclidean distance between $P$ and $Q$.

Then, we utilize the $\bm{DR}$ to compensate for the distance loss after finding the top m premade images with highest key point similarity.
By measuring with $\bm{DR}$, the image with $\bm{DR}$ closest to 1 chosen from the $m$ images with highest similarity is selected as the most similar one, and the RP tagged with this image is returned as the current user's position.

\subsection{AR Display}

After locating the users, we conduct the AR display of various indoor environment information in two steps: scene mapping and pose estimation.
Compared with most vision-based methods, our method doesn't require users to put their phone camera within a near distance ($0.5-2$ meters on average). And it does not need to extract the patterns of observed objects (like 3D model or image key points) beforehand to enable AR tracking.

\textbf{Scene mapping.}
Scene mapping is a process which maps the positions of the physical space to virtual space to enable location-based AR. The RP which can be regarded as the discretized position of the current user can be determined through two-step positioning.

The virtual space is presented by a 3D spatial coordinate system constructed from the 3D indoor map. By setting the center of the 3D indoor map as the origin, the RPs corresponding with the physical positions can be represented as coordinates in the virtual space (the directions of $x$ and $z$ axes representing right and forward while $y$ axis representing vertical). There is a size ratio between the virtual space and physical space. For simplicity, we adjust the scale of coordinates to make the size ratio equal to one. Through two-stage positioning, the RP closest to users can be chosen, and the coordinates in virtual space can be determined.

After mapping the physical position to a position in virtual space, we insert virtual objects into the virtual space. The coordinates of the virtual objects can be easily computed within virtual space. In our experiment, we add 3D texts near the tables where different projects are placed. The contents of 3D texts are the names of the projects. Note that while only 3D texts are shown in the virtual space, any 3D objects can be easily integrated similarly like the 3D texts.

\textbf{Pose estimation.}
In virtual space, the coordinate of the user (which is the coordinate of the closest RP) as well as the coordinates of all the virtual objects placed in can be determined, so the relative orientation and distance can also be computed. In order to show AR information based on the user's location, a normalized vector $f$ representing the face direction is calculated through the built-in compass and accelerator. By adding a constant $h$ representing the body height to the y coordinate of the RP, the coordinate of the eyes $e$ of the current user is derived. Then coordinates of $e$ and direction of vector $f$ are used to describe the pose of the user's eyes.

When position and orientation of the user's eyes are determined, a pyramid representing the scope of view is calculated based on the pixel aspect ratio of the camera lens, vector $f$ and the max observe distance beyond which virtual objects will not be displayed.

\begin{figure}[ptb]
\centering
\includegraphics[width=8.5cm]{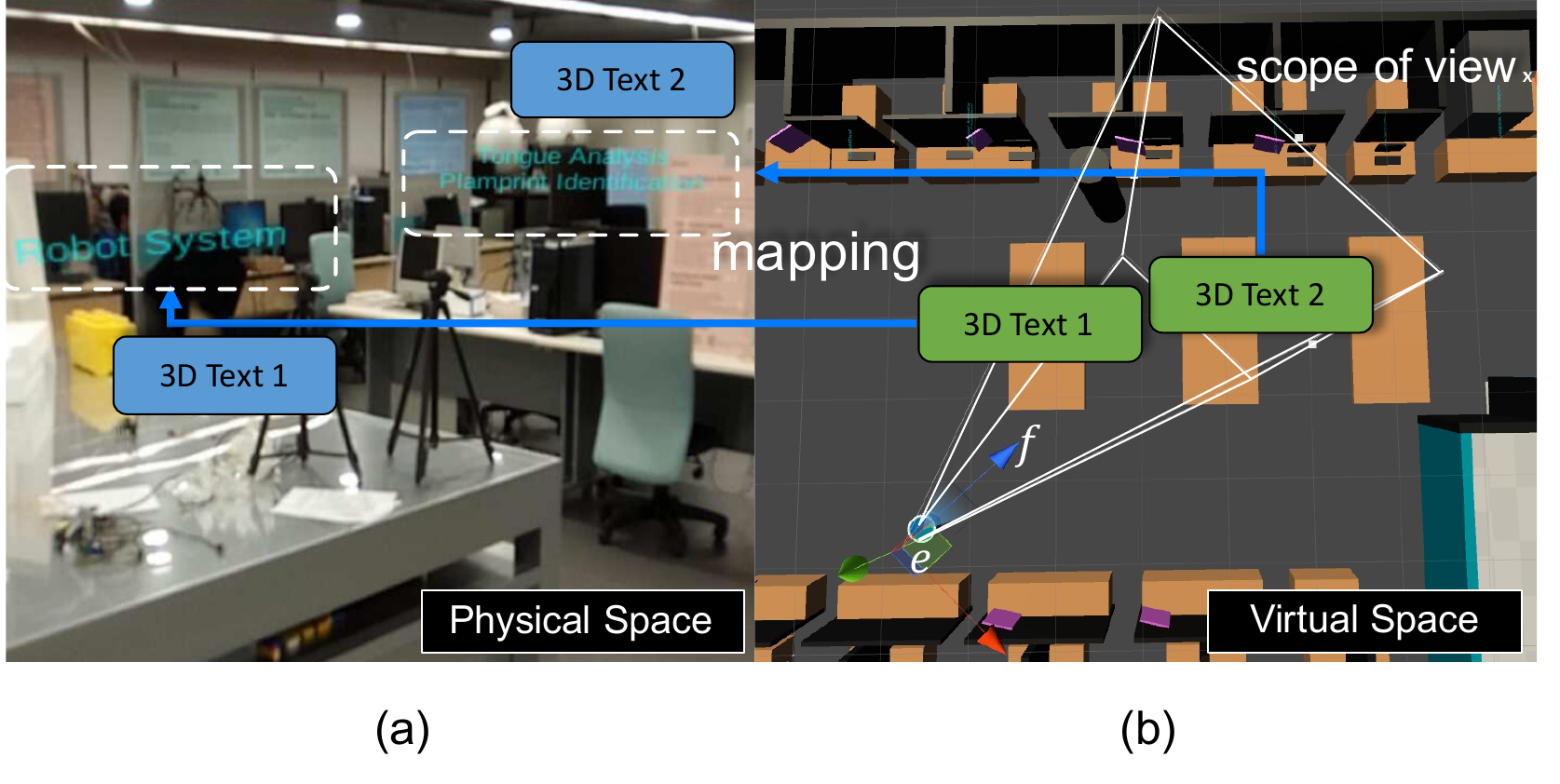}
\caption{The augmented view which implemented using the proposed method. \textbf{(a)} The view users see. \textbf{(b)} The corresponding pose of users in 3D virtual space.}
\label{fig_5}
\end{figure}

The scope of view is used in the next step to choose which virtual objects will be displayed, and the vector $f$ is used to estimate the relative angle between the users. By setting up a local Cartesian coordinate system $L$ with $e$ being the origin and $f$ being the axis with zero degree, the poses of virtual objects can be expressed in the Cartesian coordinate form. The pose information of the virtual objects in physical space can be obtained by mapping Cartesian coordinates relative to $L$ back to physical space. This process can be illustrated in Fig. \ref{fig_5}. Sub-figure (b) shows the virtual space. In (b), vector $f$ representing the user's orientation is denoted as the blue arrow , and the scope of view is shown as the pyramid in white lines. The blue texts are virtual objects put in the virtual space, while are also rendered in the physical space with relative orientations and distances in (a).

\section{Experiment Results}
\label{section4}

The experiment site is an exhibition hall. We used three types of smartphones in our experiments--Qihoo N4S, Huawei Mate 9 and Samsung Galaxy S6. Wi-Fi fingerprints are collected with these three phones randomly on all the 45 RPs on different days in a week. Photos captured at each RP are taken using these three smartphones in the offline process, and the original photos are all compressed with resolution 800*600 before feature extraction. In the online process, the photos are automatically taken and uploaded to the server by an android app we developed for every 2 seconds, and different methods are used on the photos to evaluate. We compared our method with other three different methods, including only utilizing Wi-Fi fingerprints, only utilizing key point matching, and combing Wi-Fi fingerprints with key point matching.
In order to avoid noise, the results are averaged on 300 samples collected in corresponding smartphones.

Since the alignment and trigger of virtual objects are both decided by the relative poses of these objects, so to these objects the pose of eyes relative to these virtual objects, which is also relative to the virtual space, is the only factor to affect the performance of the indoor AR in the proposed system. While the directional vector $f$ is dependent on the built-in compass and accelerator, we use the localization precision to evaluate the indoor AR performance. In this experiment, RP matching rate and average distance error are used as metrics to evaluate and compare.

\begin{figure}
\centering
\includegraphics[height=7cm ,width=8.5cm]{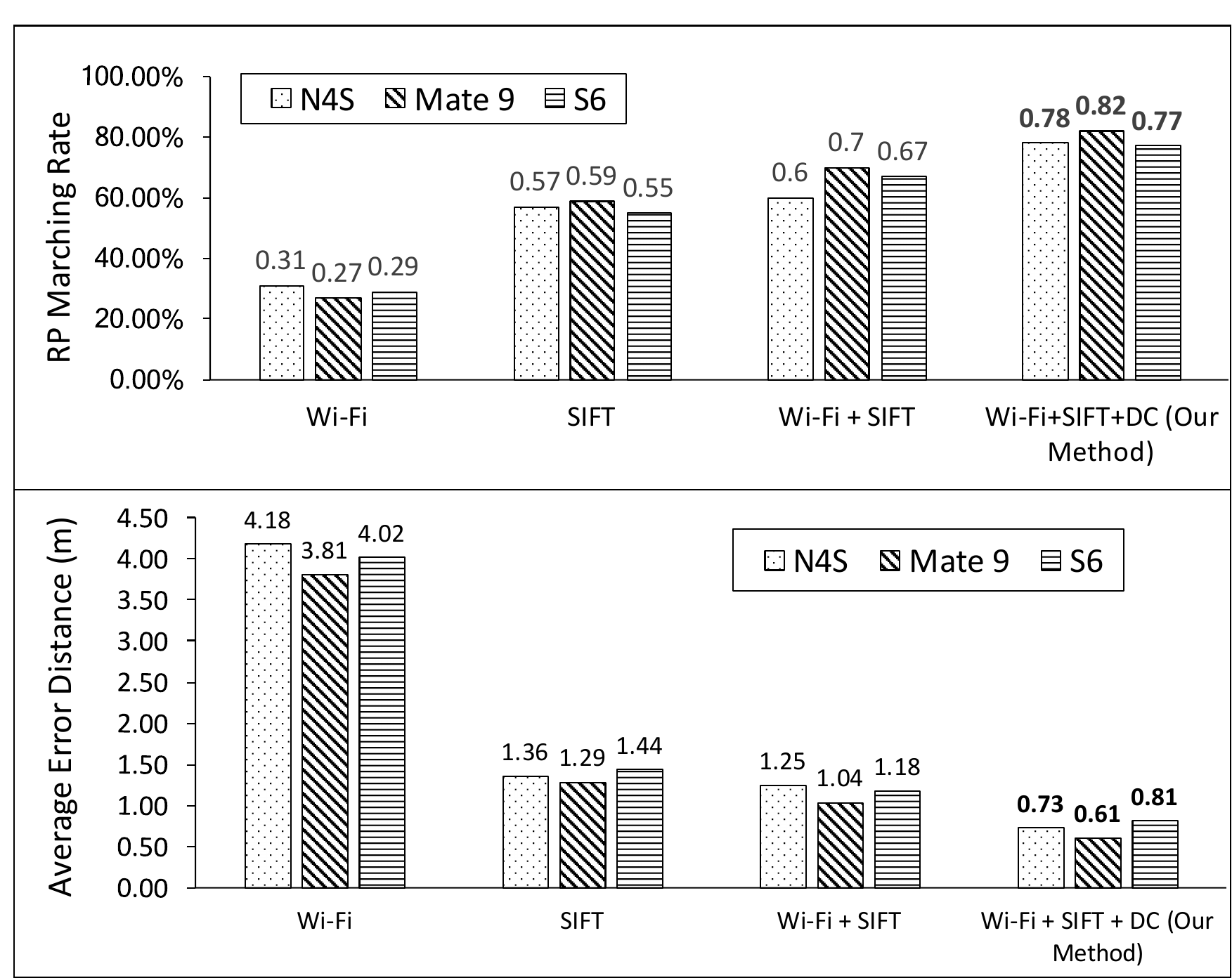}
\caption{Performance of localization accuracy(RP matching rates and average error distances of 3 phones using different methods)}
\label{fig_6}
\end{figure}

The result of RP matching rate is shown in the top diagram of Fig. \ref{fig_6}.
By evaluating the matching rate, we find indoor localization purely depending on Wi-Fi fingerprints is not accurate since Wi-Fi signals can be affected by a large amount of factors and RSS can be distinct for different kinds of smartphones. Image matching using SIFT is much better, and these three phones have similar matching rates. Combining these two methods, the matching rates increase for all three smartphones. It is a good evidence that through clustering, some incorrect images are filtered out before matching. The integration of distance compensation largely improves the matching rates, which shows the distance compensation makes closer images in a group of similar images more identifiable.

The result of average distance error is shown in bottom diagram of Fig. \ref{fig_6}.
For a mismatch, the predicted RP is relatively better when it is closer to the correct RP. So we also evaluate the precision of the proposed algorithm by calculating the average distance between the predicted RPs and the correct RPs, since matching rate alone cannot reveal the distance precision. The average error in distance is also measured under different conditions. The average error in distance of each method is shown on the top of each corresponding bar. By analyzing the distance error, it can be shown that the precision is within 1 meter, and most of the time (over 80\%) the correct RP can be chosen. It is feasible to be used to integrate with the location based system proposed in this article.

We utilized CNN as a feature extractor to replace SIFT during the RP positioning step. Features derived from ResNet-152\cite{he2016deep}, DenseNet-121\cite{huang2017densely}, and ConvNeXt-V2\cite{liu2022convnet} were compared with those in the database. Due to the nature of the generated features, distance compensation was omitted for these methods. DenseNet-121 achieved the best result with an average error distance of approximately 1.3, while the other two methods exhibited average error distances exceeding 1.6, performing less effectively than our proposed approach. This experiment demonstrates that while the proposed method, based on SIFT and distance compensation, is simpler in nature, it achieves superior precision.

% Further experiments are conducted on the performance of the proposed indoor localization algorithm. Our proposed method can achieve higher precision compared with methods from Torri\cite{torii2011visual} in the test environment. The comparison is shown in Fig. \ref{fig_7}.

% \begin{figure}
% \centering
% \includegraphics[height=6cm ,width=8.5cm]{pic/LOC3.png}
% \caption{The comparison of indoor localization precision}
% \label{fig_7}
% \end{figure}

\begin{figure}
\centering
\includegraphics[height=3.8cm ,width=8.5cm]{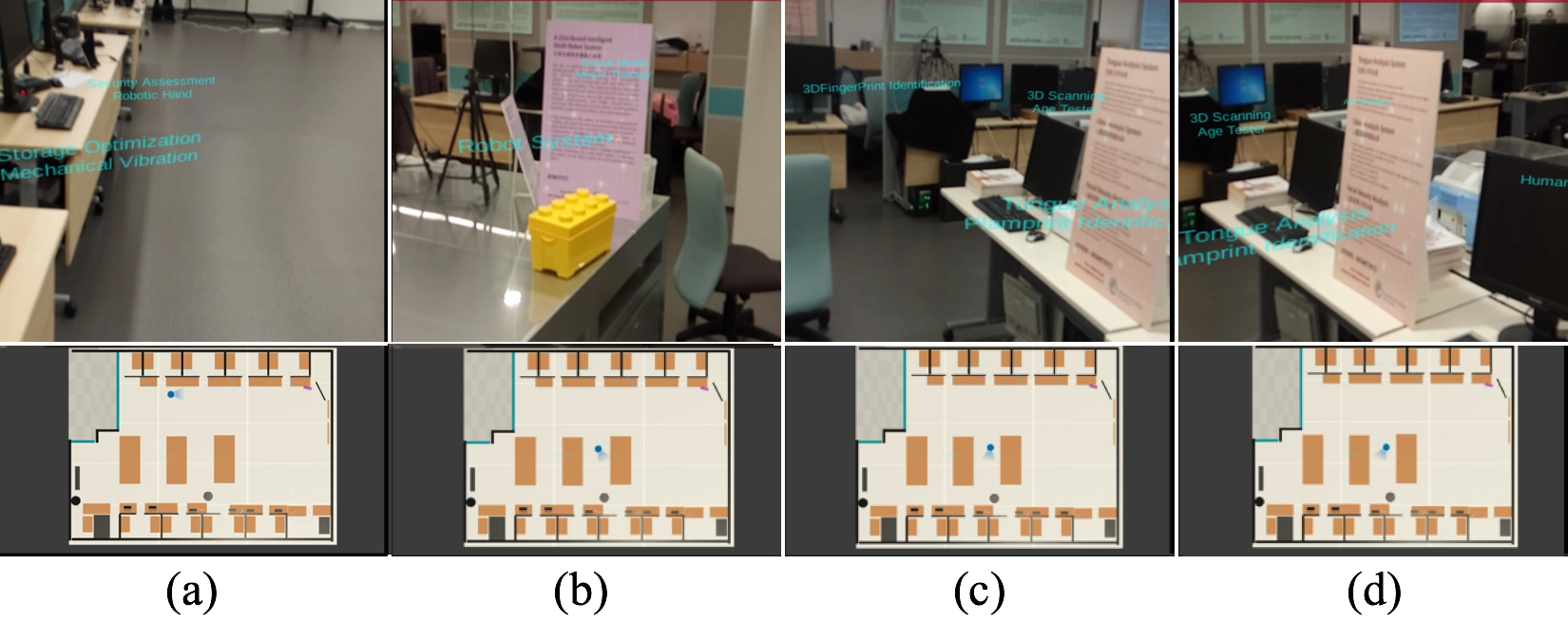}
\caption{The screenshots of the android app with proposed indoor AR implementation. \textbf{Row 1}: screenshots of the augmented scenes with virtual texts. \textbf{Row 2}: the positions and orientations of the users corresponding to the Row 1 view}
\label{fig_8}
\end{figure}

To test the actual performance of the indoor AR system, the tester using the android app implemented with the proposed method in the exhibition hall. The screenshots of the AR application on Qihoo N4S are shown on the first row of Fig. \ref{fig_8}. The blue augmented 3D texts are overlaid before the exhibition booths on the real scenes, which indicate the project names of booth in front of the tester. The second row of Fig. \ref{fig_8} displays the tester's current position and orientation in the hall. The 2D projection of the 3D map now serves as the indoor map.

\section{Conclusion}
\label{section5}

In the increasing pursuit of indoor localization precision and public fever on AR, we believe indoor AR techniques will become even more widely demanded than ever. Although there are many types of implementations in both indoor localization and AR and some combination with these two, we assume the most promising implementations should have these features. First, they should be low in cost, both for deployment and maintenance. Second, they should be robust, resistant to ambient interference. And they can be easily fitted in various scenarios as a service. Unlike most implementations using extra hardware like ibeacons and RFID, the indoor AR framework proposed in this paper doesn't need any extra hardware other than users' ordinary smartphones. And compared with most systems using pure RSS, the camera as well as the built-in sensors like compass and accelerator is used to reduce occasional error in our framework. To further make the whole framework more accurate, image matching using SIFT is further optimized with distance compensation. To solve the limitations on images in vision-based AR, such as intrusive markers and limited observing distance, a new implementation based on relative Cartesian coordinates is also proposed.

Although the performance of this framework is strongly related to the value of sampling grid intervals, the selections of interval values can be flexible and customized in different small areas based on size of environments and requirements of AR objects. The major online operations, such as subarea positioning and RP positioning with distance compensation, are processed on servers without heavy computation cost (like CNN). And these methods all support parallel computation on several cores or slave servers(nodes) since the implementation only contains simple iterations without inter-dependencies. So, our framework can meet the requirements of most weak real-time scenarios while remaining scalable in large environments.

\section*{Acknowledgments}

We would like to express our gratitude to WPI, Tohoku University and Purdue University, which provided the facilities and support necessary for the successful completion of this research. Special thanks to Haichong (Kai) Zhang, Ziming Zhang, Yamada K.D, and Emma Ge for their invaluable guidance, insightful comments, and constructive feedback throughout the development of this project.

We also acknowledge the contributions of our experiment participants, whose involvement was crucial in validating our method. Additionally, we appreciate the support of our families and friends for their encouragement and understanding during the research process.

Finally, we would like to extend our gratitude to the anonymous reviewers for their time and effort in reviewing this manuscript. 
%
% \smallskip
% \begin{quote}
% \small
% \begin{verbatim}
% \section*{Acknowledgments}
% \end{verbatim}
% \end{quote}
% \smallskip
%
% This must be set immediately following the last numbered section of the paper.

% To test the citation, see \cite{rf1,rf2,rf3,rf4,rf5,rf6,rf7,rf8}.

% \bibliographystyle{iisnum}
% \bibliography{BibSample}
\bibliographystyle{IEEEtran}
\bibliography{hsl}

\end{document}